\documentclass[10pt,twocolumn,letterpaper]{article}

\usepackage[pagenumbers]{cvpr} % To force page numbers, e.g. for an arXiv version

\usepackage{booktabs} % For professional looking tables
\usepackage{multirow} % For multirow cells
\usepackage{amsmath}
\usepackage{amssymb}
\usepackage{graphicx}
\usepackage{stfloats}
\usepackage{algorithm}
\usepackage{algorithmicx}
\usepackage{algpseudocode}
\usepackage{colortbl}
\definecolor{cvprblue}{rgb}{0.21,0.49,0.74}
\usepackage[pagebackref,breaklinks,colorlinks,allcolors=cvprblue]{hyperref}

\def\paperID{13364} % *** Enter the Paper ID here
\def\confName{CVPR}
\def\confYear{2025}

\title{Mogo: RQ Hierarchical Causal Transformer for High-Quality 3D Human Motion Generation}

\author{Dongjie Fu\\
Mogo AI\\
{\tt\small mirecofu@163.com}
}

\begin{document}

\twocolumn[{
\renewcommand\twocolumn[1][]{#1}
\maketitle
% \vspace{-3mm}
\begin{center}
    \centering
    \captionsetup{type=figure}
    \includegraphics[width=0.95\linewidth]{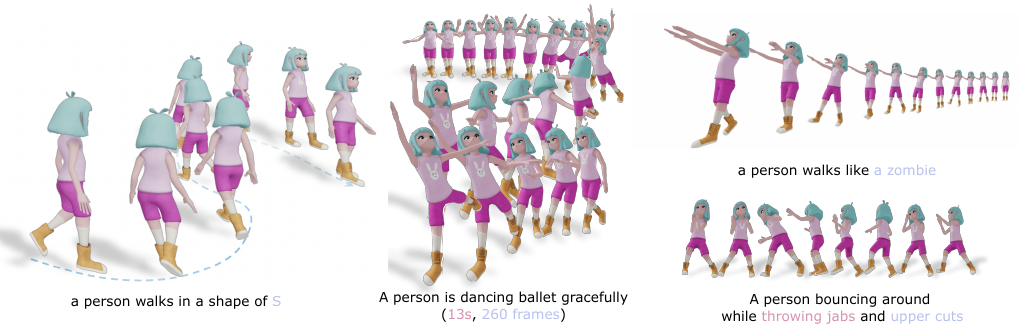}
        	 % \vspace{-1.5mm}

    \captionof{figure}{Mogo is a novel GPT-type model for high-quality text-to-motion generation, capable of producing longer sequences and handling open-vocabulary prompts. Given inputs \eg \textit{"A person is dancing ballet gracefully", "A person walks like a zombie"}. Mogo generates vivid, lifelike 3D human motions at once.}
    \label{fig:main_visual}
    	 % \vspace{-1.5mm}
\end{center}
}]
% \begin{figure*}[!t] % 使用 figure 环境来处理图片
%     \centering
%     % 按照屏幕宽度自适应，高度保持比例填充
%     \includegraphics[width=\textwidth]{main-figure.pdf}
%     \caption{The architecture of the model inference} % 添加图片标题
%     \label{fig:inference} % 添加图片标签
% \end{figure*}
\begin{abstract}
    In the field of text-to-motion generation, Bert-type Masked Models (\eg MoMask, MMM) currently produce higher-quality outputs compared to GPT-type autoregressive models (\eg T2M-GPT). However, these Bert-type models often lack the streaming output capability required for applications in video game and multimedia environments, a feature inherent to GPT-type models. Additionally, they demonstrate weaker performance in out-of-distribution generation.
    To surpass the quality of BERT-type models while leveraging a GPT-type structure—without adding extra refinement models that complicate scaling data, we propose a novel architecture, Mogo (Motion Only Generate Once), which generates high-quality lifelike 3D human motions by training a single transformer model.
    Mogo consists of only two main components: 1) \textbf{RVQ-VAE}, a hierarchical residual vector quantization variational autoencoder, which discretizes continuous motion sequences with high precision; 2) \textbf{Hierarchical Causal Transformer}, responsible for generating the base motion sequences in an autoregressive manner while simultaneously inferring residuals across different layers. 
    Experimental results demonstrate that Mogo can generate continuous and cyclic motion sequences up to 260 frames (13 seconds), surpassing the 196 frames (10 seconds) length limitation of existing datasets like HumanML3D. On the HumanML3D test set, Mogo achieves a Fréchet Inception Distance (FID) score of 0.079, outperforming both the GPT-type model T2M-GPT (FID = 0.116), AttT2M (FID = 0.112) and the BERT-type model MMM (FID = 0.080). Furthermore, our model achieves the best quantitative performance in out-of-distribution generation.
\end{abstract}    
\section{Introduction}
\label{sec:intro}

In recent years, the study of generating 3D human motion from textual descriptions has gained increasing attention in fields such as video games, VR/AR, animations, and humanoid robotics, leading to extensive research efforts~\cite{guo2023momask,zhang2023generating,zhang2022motiondiffuse,zhang2023remodiffuse,pinyoanuntapong2024mmmgenerativemaskedmotion,8885540,wang2019learningdiversestochastichumanaction}. Among these studies, models using transformer~\cite{Vaswani2017AttentionIA} algorithms have demonstrated significant advantages in both generation quality and real-time performance~\cite{guo2023momask,zhang2023generating, pinyoanuntapong2024mmmgenerativemaskedmotion,Zhong_2023_ICCV}. Typically, transformer-based motion generation models consist of two components: 1) a vector quantized variational autoencoder (VQ-VAE) to discretize continuous 3D human motion into quantized tokens, and 2) a transformer-based model trained to generate these discrete motion tokens through conditional reasoning. Despite notable improvements over previous approaches, these methods still exhibit inherent limitations: VQ quantization inevitably introduces some errors due to encoding granularity, impacting the overall generation quality. Moreover, autoregressive transformer models suffer from long-range attention loss when generating long sequences. 

To address these limitations, AttT2M~\cite{Zhong_2023_ICCV} incorporated a body-part attention-based spatio-temporal feature extraction method(BPST), to more finely train the VQ quantization model, while MoMask adopts RVQ-VAE, a residual vector quantization autoencoder, to achieve higher-quality motion quantization. To overcome the limitations of unidirectional autoregressive transformer models in generating high-quality motion, models like MoMask~\cite{guo2023momask} and MMM~\cite{pinyoanuntapong2024mmmgenerativemaskedmotion} employ a masked transformer encoder architecture to deliver superior generation results. 

However, these impressive results still leave certain issues unresolved. Fine-grained VQ quantization models~\cite{Zhong_2023_ICCV} require detailed motion joints splitting during the encoder training phase, but it did not significantly improve the generation quality of T2M-GPT (FID increased from 0.116 to 0.112). Bert-type models lose the token-by-token streaming output capability required for low-latency frame-by-frame generation in video games and multimedia environments, and they often exhibit a higher risk of overfitting along with reduced out-of-distribution generation capability. Models using RVQ-VAE as the encoder require two models to be trained during the generation phase~\cite{guo2023momask,Chang2023MuseTG}: one model (M-Transformer) generates base motion sequences based on the codebook's base layer, while another model (R-Transformer) generates the motion residuals from the remaining codebook layers. This design introduces additional model training overhead when scaling data. 
Given the success of GPT-type generative models in various domains, as well as research indicating their superiority in few-shot learning, scaling effects, long-sequence generation, and multimodal applications~\cite{brown2020languagemodelsfewshotlearners,kaplan2020scalinglawsneurallanguage,Xu2022MultimodalLW,Wang2024Emu3NP}, we hypothesize that GPT-type models are well-suited for motion generation, which inherently requires long-sequence generation and multi-style adaptation. 

In light of these insights, we propose an innovative architecture named \textbf{Mogo}. By employing RVQ-VAE as a high-quality motion tokenizer and utilizing a single hierarchical causal transformer model for inference, Mogo achieves superior generation quality and generalization capability over BERT-type models. Our contributions are as follows:

\begin{enumerate}
  \item We designed a simple model structure that requires only a hierarchical GPT-type model (without the refine model used in \cite{guo2023momask}) to apply the codebook information generated by the RVQ-VAE encoder across all layers. This design significantly facilitates future efforts toward streaming output and scaling data for the model.
  
  \item Mogo significantly enhances the generation quality of GPT-type motion generation models. The metrics achieve state-of-the-art (SOTA) performance among GPT-type models on the HumanML3D \cite{Guo_2022_CVPR} and KIT-ML \cite{Plappert2016} test sets. On the CMP\cite{CombatMotion} test set zero-shot inference evaluation, Mogo reaches SOTA level for all transformer-based models.

  \item The generation length of Mogo breaks through the limitations of the training dataset's motion sequence length. For instance, the maximum motion data sequence length in HumanML3D \cite{Guo_2022_CVPR} is 10 seconds (196 frames), while our trained model can generate continuous and cyclic motion sequences of up to 13 seconds(260 frames).
  
  \item We optimized the input prompts during inference through the LLM model to enhance Mogo's generation performance in zero-shot and few-shot scenarios.
\end{enumerate}

\section{Related work}
\label{sec:relate_work}
\begin{figure*}[htbp] % 使用 figure 环境来处理图片
    \centering
    % 按照屏幕宽度自适应，高度保持比例填充
    \includegraphics[width=\textwidth]{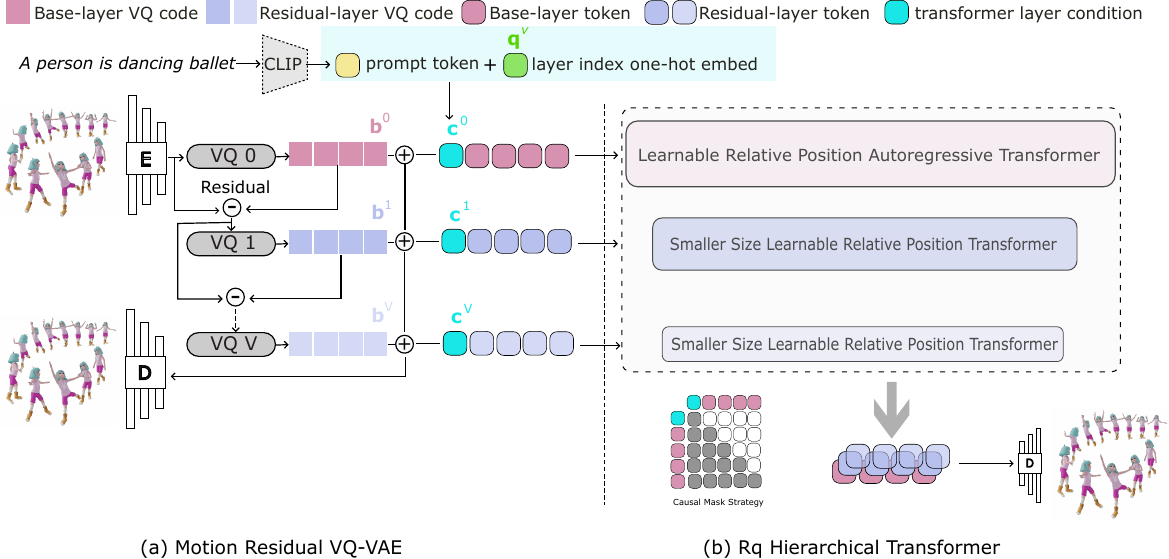}
    \caption{The architecture of the model:(a) \textbf{RVQ-VAE} is a hierarchical residual vector quantization variational autoencoder that discretizes continuous motion sequences with high precision; (b) a \textbf{Rq Hierarchical Causal Transformer} generates base motion sequences autoregressively while inferring residuals across layers.
    } % 添加图片标题
    \label{fig:architechure1} % 添加图片标签
\end{figure*}
\textbf{Human Motion Generation.}
Human motion generation based on different conditions, such as text, audio, music, or image inputs, has made significant progress in recent years \cite{zhu2023humanmotiongenerationsurvey}. Early works\cite{8885540, Ghosh_2021_ICCV} commonly model motion generation deterministically. However, this type of model often results in vague and uncertain motions during generation. Stochastic models can effectively address this issue. \cite{Cai_2018_ECCV, wang2019learningdiversestochastichumanaction} employed GAN models to generate motion sequences based on conditions. \cite{guo2022action2video, petrovich21actor} used temporal VAE and transformer architecture to model and infer motions. In the field of text-based motion generation, \cite{Guo_2022_CVPR} used temporal VAE to model the probabilistic mapping between text and motions. With the widespread application of diffusion \cite{rombach2021highresolution} and transformer \cite{Vaswani2017AttentionIA} architecture in text and image domains, their potential in motion generation has gradually been explored. Works such as \cite{kim2022flame, zhang2022motiondiffuse, zhang2023remodiffuse, kong2023prioritycentrichumanmotiongeneration, chen2023executing, tevet2023human} utilized diffusion architecture to train models for motion generation. \cite{guo2023momask, pinyoanuntapong2024mmmgenerativemaskedmotion} adopted a Bert-type structure in their model designs. \cite{zhang2023generating, Zhong_2023_ICCV, jiang2024motiongpt} used GPT-type autoregressive transformer models for text-to-motion generation. Among them, \cite{jiang2024motiongpt} achieved multimodal input-output for text-to-motion and motion-to-text by fine-tuning a language model.
\\
\textbf{GPT-type Models.}
We categorize motion generation models with decoder-only transformers as GPT-type models, and those using masked token bidirectional attention as BERT-type models. Current research indicates that while BERT-type models slightly outperform GPT-type models on generation quality metrics, GPT-type models have shown distinct advantages in language modeling, excelling in few-shot learning, scalability, and long-sequence generation \cite{brown2020languagemodelsfewshotlearners, kaplan2020scalinglawsneurallanguage, Wang2024Emu3NP}. As motion generation datasets grow and application scopes broaden, GPT-type models are expected to reveal greater potential. Experiments by \cite{zhang2023generating} confirm that GPT-type models continue to benefit from scaling effects in motion generation. AttT2M \cite{Zhong_2023_ICCV} addresses accuracy limitations of GPT models, reducing the FID from 0.116 in T2M-GPT \cite{zhang2023generating} to 0.112 on the HumanML3D test set \cite{Guo_2022_CVPR} by refining body joints segmentation encoding. MotionGPT \cite{jiang2024motiongpt}, trained with a large language model \cite{2020t5}, supports bidirectional input and output of motion and text descriptions but achieves an FID of 0.232.\\
\textbf{Hierarchical Transformer Model.}
In solving complex problems, hierarchical transformer models have been widely applied in natural language processing \cite{nawrot-etal-2022-hierarchical, Pappagari2019HierarchicalTF}, image generation \cite{ding2022cogview2, nawrot-etal-2022-hierarchical}, and computer vision \cite{liu2021Swin, chen2022scaling}, demonstrating significant advantages. Unlike traditional flat structures, hierarchical structures allow the model to capture fine-grained features at different levels, thus more efficiently handling high-dimensional and more complex tasks. For example, \cite{nawrot-etal-2022-hierarchical, Pappagari2019HierarchicalTF} significantly enhanced the model's learning and reasoning capabilities for long sequences using hierarchical structures. \cite{chen2022scaling} successfully expanded the processing resolution of computer vision to 4K levels using hierarchical transformer models, showcasing their strong capabilities in handling large-scale visual data. Similarly, \cite{ding2022cogview2} achieved state-of-the-art performance in image generation using hierarchical transformer models, proving the effectiveness of this structure in complex generation tasks.
\\
\textbf{Motion Tokenizer.}
TM2T \cite{chuan2022tm2t} first introduced VQ-VAE into the field of motion generation, mapping continuous human motions to discrete tokens. T2M-GPT \cite{zhang2023generating} further optimized the performance of VQ-VAE through EMA and codebook reset techniques. However, quantization errors still exist during motion reconstruction. AttT2M \cite{Zhong_2023_ICCV} reduced quantization errors during the quantization phase by performing fine-grained segmentation of human body motions. MoMask \cite{guo2023momask} not only generates base motion sequence tokens using RVQ-VAE but also captures tokens that represent residual information, significantly improving the reconstruction accuracy of discrete motion sequences.

\section{Methods}
\label{sec:methods}
\begin{figure*}[htbp] % 使用 figure 环境来处理图片
    \centering
    % 按照屏幕宽度自适应，高度保持比例填充
    \includegraphics[width=0.9\textwidth]{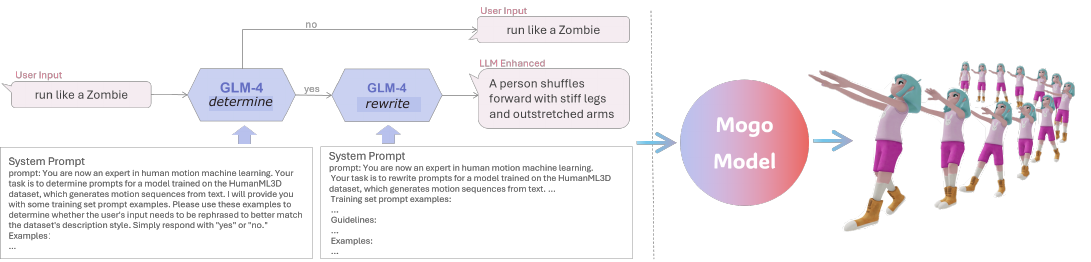}
    \caption{The architecture of prompt engineering for enhanced model inference} % 添加图片标题
    \label{fig:inference} % 添加图片标签
\end{figure*}
Our goal is to generate a high-quality motion sequence $\mathbf{m}_{1:N}$ of length $N$ from a textual description $c$, where $\mathbf{m}_i \in \mathbb{R}^D$ and $D$ represents the dimensionality of the motion pose. As illustrated in \cref{fig:architechure1}, our model architecture consists of two core components: a residual quantization-based encoder for discretizing the motion sequence into multiple layers of motion tokens (\cref{sec:trainRQVAE}), and a Hierarchical Causal Transformer model for inferring and generating these multi-layer motion tokens in a single pass (\cref{sec:trainHMT}),The prompt engineering for our inference process is presented in \cref{sec:promptEng}.

\subsection{Training: Motion Residual VQ-VAE}
\label{sec:trainRQVAE}
We largely base our RVQ-VAE design on MoMask \cite{guo2023momask}, but we introduce some modifications and optimizations in the sampling strategy. The Residual Quantizer (RQ) expresses a motion latent sequence \(\tilde{\mathbf{b}}\) as an ordered sequence of \(V+1\) codes using \(V+1\) quantization layers. We denote this as:
\({\mathcal{RQ}(\tilde{\mathbf{b}}_{1:n}) = [\mathbf{b}_{1:n}^v]_{v=0}^V}\). 
where \( b_{1:n}^v \in \mathbb{R}^{n \times d} \) represents the code sequence at the \(v\)-th quantization layer. The residual value at layer 0 is \(\mathbf{r}^0 = \tilde{\mathbf{b}}\), and the sequences \(\mathbf{b}^v\) and \(\mathbf{r}^{v+1}\) at subsequent layers are calculated as:

\begin{equation}
\mathbf{b}^v = \mathcal{Q}(\mathbf{r}^v), \quad \mathbf{r}^{v+1} = \mathbf{r}^v - \mathbf{b}^v,
\end{equation}
\\
After RQ computation, the final approximation of the latent sequence \(\tilde{\mathbf{b}}\) is the sum of all quantized sequences across layers: \(\sum_{v=0}^{V}\mathbf{b}^{v}\), which serves as input to the decoder \({D}\) for motion reconstruction.
\\
\textbf{Training Loss Function.} The loss function for training the RVQ-VAE model is defined as follows:
\begin{equation}
    \mathcal{L}_{rvq} = \|\mathbf{m} - \hat{\mathbf{m}}\|_1 + \beta \sum_{v=1}^{V}\|\mathbf{r}^v - \mathrm{sg}[\mathbf{b}^v]\|_2^2,
\end{equation}
where \(\mathrm{sg}[$·$]\) denotes the stop-gradient operation, and \(\beta\) is a hyperparameter controlling the commitment loss. This loss is optimized using the straight-through gradient estimator \cite{10.5555/3295222.3295378}, and as in \cite{guo2023momask,zhang2023generating}, we employ codebook resetting and exponential moving average to update the codebook. After training, each motion sequence \(\mathbf{m}\) can be represented by \(V+1\) discrete token sequences \(T = [t_{v}^{1:n}]_{v=0}^{V}\), where each token sequence \(t_{v}^{1:n} \in \{1, \ldots, |\mathcal{C}_{v}|\}^{n}\) is an ordered codebook index of the quantized embeddings \(\mathbf{b}_{v}^{1:n}\), such that \(\mathbf{b}_{i} = \mathcal{C}_{v} t_{v}^{i}\) for \(i \in [1, n]\). Among these \(V+1\) sequences, the first (base) sequence contains the most significant information, while the subsequent layers gradually contribute less.
\\
\textbf{Quantization Optimization Strategy.}
Although our RVQ-VAE adopts a similar quantization approach as \cite{guo2023momask,zhang2023generating}, we make improvements in the sampling strategy. Instead of using convolutions with a stride of 2 and nearest-neighbor interpolation for downsampling and upsampling, as in \cite{guo2023momask,zhang2023generating}, we adjust the convolution stride to 1. This enhances the precision and expressiveness of motion reconstruction, enabling finer feature extraction and smoother reconstruction processes.

\subsection{Training: Hierarchical Causal Transformer}
\label{sec:trainHMT}
\textbf{Hierarchical Architecture Design.} After encoding the motion sequences using RQ-VAE, we obtain \(V+1\) discrete motion token sequences for each motion sequence. To handle these layers in a unified manner, we designed a hierarchical Transformer model capable of processing the features from each layer simultaneously. For each layer of the Transformer model, we construct the input $\mathbf{s}^v$, which includes the text embedding, the quantization layer embedding, and the summed representations of the motion embeddings from the previous $v$ layers:
\begin{equation}
    \mathbf{s}^v = [\mathbf{p} + \mathbf{q}_{\text{emb}}, \mathbf{t}_{v}^{1:n}]
\end{equation}
Here, $\mathbf{p}$ represents the text prompt sentence embedding obtained from the pretrained CLIP model \cite{Radford2021LearningTV}, capturing the global relationship between the sentence and the motion. $\mathbf{q}_{\text{emb}}$ is the embedding of the quantization layer index $v$, computed as: $\mathbf{q}_{\text{emb}} = \mathbf{W} \cdot \mathbf{Q}_{\text{one-hot}}$, where \(\mathbf{W} \in \mathbb{R}^{m \times n}\) is the weight matrix representing the linear transformation of the quantization embedding, and \(\mathbf{Q}_{\text{one-hot}} = \text{one-hot}(\mathbf{qid})\) is the one-hot representation of the layer index $\mathbf{qid}$. We sum $\mathbf{p}$ and $\mathbf{q}_{\text{emb}}$ to obtain the mixed motion generation condition $\mathbf{c}$  (PnQ condition), which serves as the prefix condition for the input sequence at each transformer layer.
$\mathbf{t}_{v}^{1:n} = \sum_{i=0}^{v} \mathbf{t}_{i}^{1:n}, \quad \forall v \in \{1, 2, \dots, V\}$ represents the features of the current layer $v$, obtained by summing the embedded features of the ordered codebook index sequences $t_{v}^{1:n}$ from all previous layers up to layer $v$. This design allows for a cumulative construction, ensuring that the features of the current layer are built upon the accumulated information from all previous layers and the layer-wise sequential information.

We also tested input sequences without mixed conditions: \(\mathbf{s}^v = [\mathbf{p},\mathbf{q}_{\text{emb}}, \mathbf{t}_{v}^{1:n}]\), with results detailed in \cref{sec:ablation}.
\\
\textbf{Relative Positional Encodings.} As described in \cref{sec:trainRQVAE}, we adjusted the RQ-VAE convolution stride from \(2\) to \(1\) \cite{guo2023momask,zhang2023generating}, resulting in a motion sequence length that is four times longer after encoding. To address the challenges posed by this increase in sequence length, we were inspired by Transformer-XL \cite{dai-etal-2019-transformer} and incorporated the relative positional encoding attention architecture introduced in \cite{dai-etal-2019-transformer}. Original attention mechanisms using absolute positional encoding often suffer from performance degradation when handling longer sequences, especially when the sequence length during inference differs from that during training \cite{dai-etal-2019-transformer}. By adopting the relative positional encoding, we not only enhanced the model’s ability to capture long-range attention dependencies but also improved the coherence and naturalness of the generated motion sequences.

The attention computation with relative positional encoding is defined as follows:
    
\begin{equation}
    \begin{aligned}
        \mathbf{A}_{i,j}^{\text{rel}} = & \underbrace{\mathbf{E}_{x_i}^\top \mathbf{W}_q^\top \mathbf{W}_{k,E} \mathbf{E}_{x_j}}_{(a)} + \underbrace{\mathbf{E}_{x_i}^\top \mathbf{W}_q^\top \mathbf{W}_{k,R} \mathbf{R}_{i-j}}_{(b)} \\
    & + \underbrace{u^\top \mathbf{W}_{k,E} \mathbf{E}_{x_j}}_{(c)} + \underbrace{v^\top \mathbf{W}_{k,R} \mathbf{R}_{i-j}}_{(d)}
    \end{aligned}
\end{equation}

In this formulation, ${(a)}$ \(\mathbf{E}_{x_i}^\top \mathbf{W}_q^\top \mathbf{W}_{k,E} \mathbf{E}_{x_j}\) represents the linear transformation of the embedding \(\mathbf{E}_{x_j}\) at position \(j\) via the embedding vector \(\mathbf{E}_{x_i}\), query weights \(\mathbf{W}_q\), and key weights \(\mathbf{W}_{k,E}\), capturing the direct influence of position \(i\) on position \(j\). ${(b)}$ \(\mathbf{E}_{x_i}^\top \mathbf{W}_q^\top \mathbf{W}_{k,R} \mathbf{R}_{i-j}\) calculates the relationship between position \(i\) and the relative position \(\mathbf{R}_{i-j}\), emphasizing their distance. ${(c)}$ \(u^\top \mathbf{W}_{k,E} \mathbf{E}_{x_j}\) applies a linear transformation to \(\mathbf{E}_{x_j}\) with parameters \(u\) and weights \(\mathbf{W}_{k,E}\), supplementing the information associated with position \(j\). Finally, ${(d)}$ \(v^\top \mathbf{W}_{k,R} \mathbf{R}_{i-j}\) transforms the relative position \(\mathbf{R}_{i-j}\) via parameters \(v\) and weights \(\mathbf{W}_{k,R}\), reflecting the positional relationship between \(i\) and \(j\). Thus, in each layer of the Transformer, the autoregressive attention computation is expressed as:

\begin{equation} 
    \mathbf{a}_{v}^{n} = \text{Masked-Softmax}(\mathbf{A}_{v}^{n})\mathbf{V}_{v}^{n}
\end{equation}

where $\mathbf{a}_{v}^{n}$ denotes the attention output at the $n$-th layer of the Transformer model for layer $v$, $\mathbf{A}_{v}^{n}$ is the attention matrix with relative positional encoding.
\\
\textbf{Loss Function.} The goal of the model is to autoregressively generate motion tokens based on the text prompt input \(c\), aiming to make the generated sequence as close as possible to the real motion sequence. We use a maximum likelihood estimation function to compute the model's loss value, optimizing the model parameters by minimizing the log-likelihood difference between each generated token and the target token:

\begin{equation}
\mathcal{L}_{\text{ce}} = -\frac{1}{B} \sum_{i=1}^{B} \sum_{j=1}^{T_i} \sum_{v=1}^{V} \log p_\theta(t_j \mid O_{i,j-1}^{v}, c)
\end{equation}

where \(B\) denotes the batch size, \(T_i\) is the sequence length of the \(i\)-th sample, and \(V\) is the number of quantization layers. \(O_{i,j-1}^{v}\) represents the output at the \(v\)-th layer for the \(i\)-th sample when generating the \(j\)-th token, conditioned on the previous \(j-1\) generated tokens and the text prompt \(c\). This design allows the model to better integrate outputs from different levels, enhancing the quality of the generation.
\\
\textbf{Data Processing.} Following T2M-GPT \cite{zhang2023generating}, we used the same data augmentation strategy: during training, \(\tau \times 100\%\) of real code indices are replaced with random ones, where \(\tau\) is either a hyperparameter or randomly sampled from \(U[0,1]\). In our case, \(\tau\) is set to 0.5. This method improves the model's generalization capability.

\subsection{Inference Capability Optimization}
\label{sec:promptEng}
During user feedback experiments, we found that most participants provided text descriptions not present in the dataset, such as zombie, warrior, or ninja motions. While OMG \cite{liang2024omg} improves the Zero/Few-shot generation capability of diffusion models through a motion ControlNet, recent studies on prompt engineering for large language models \cite{ye-etal-2024-prompt, Brown2020LanguageMA} highlight the generative strengths of GPT in similar scenarios. Consequently, we developed a prompt engineering framework (\cref{fig:inference}) using GLM-4 \cite{glm2024chatglm} to evaluate and optimize user prompts based on HumanML3D text labels. By utilizing labeled examples, GLM-4 learns descriptive styles and rewrites user prompts to better align the generated output with user expectations.

\section{Experiments}
\label{sec:experiments}
\begin{table*}[htbp]
  \centering
  \resizebox{\textwidth}{!}{ % Adjust table to fit the page
  \begin{tabular}{c c c c c c c c}
    \hline
    \multirow{2}{*}{\textbf{Datasets}} & \multirow{2}{*}{Methods} & \multicolumn{3}{c}{R Precision${\uparrow}$} & \multirow{2}{*}{FID${\downarrow}$} & \multirow{2}{*}{MultiModal Dist${\downarrow}$} & \multirow{2}{*}{MultiModality${\uparrow}$} \\ \cline{3-5}
                                        &                                  & Top 1  & Top 2  & Top 3  &                            &                                    &                                      \\ \hline
    \multirow{9}{*}{\textbf{HumanML3D}} 
                                        
                                        & MotionDiffuse \cite{zhang2022motiondiffuse} & 0.491$^{\pm0.001}$ & 0.681$^{\pm0.001}$ & 0.782$^{\pm0.001}$ & 0.630$^{\pm0.001}$ & 3.113$^{\pm0.001}$ & 1.553$^{\pm0.042}$ \\
                                        & T2M-GPT $^{\dagger}$ \cite{zhang2023generating} & 0.491$^{\pm0.003}$ & 0.680$^{\pm0.002}$ & 0.775$^{\pm0.002}$ & 0.116$^{\pm0.004}$ & 3.118$^{\pm0.011}$ & 1.856$^{\pm0.011}$ \\
                                        & Fg-T2M\cite{wang2023fgt2mfinegrainedtextdrivenhuman}& 0.492$^{\pm0.002}$ & 0.683$^{\pm0.003}$ & 0.783$^{\pm0.003}$ & 0.243$^{\pm0.019}$ & 3.109$^{\pm0.007}$ & 1.614$^{\pm0.049}$ \\
                                        & AttT2M $^{\dagger}$\cite{Zhong_2023_ICCV} & 0.499$^{\pm0.005}$ & 0.690$^{\pm0.006}$ & 0.786$^{\pm0.004}$ & 0.112$^{\pm0.004}$ & 3.038$^{\pm0.016}$ & \textbf{2.452}$^{\pm0.043}$ \\
                                        & MotionGPT $^{\dagger}$\cite{jiang2024motiongpt} & 0.492$^{\pm0.003}$ & 0.681$^{\pm0.003}$ & 0.778$^{\pm0.002}$ & 0.232$^{\pm0.008}$ & 3.096$^{\pm0.009}$ & 2.008$^{\pm0.084}$ \\
                                        & MoMask\cite{guo2023momask} & \textbf{0.521$^{\pm0.002}$} & \textbf{0.713$^{\pm0.002}$} & \textbf{0.807$^{\pm0.002}$} & \textbf{0.045$^{\pm0.002}$} & \textbf{2.958}$^{\pm0.008}$ & 1.241$^{\pm0.040}$\\
                                        & MMM \cite{pinyoanuntapong2024mmmgenerativemaskedmotion} & 0.504$^{\pm0.002}$ & \underline{0.696}$^{\pm0.003}$ & 0.794$^{\pm0.004}$ & 0.080$^{\pm0.004}$ & \underline{2.998}$^{\pm0.007}$ & 1.226$^{\pm0.035}$ \\
                                        \cline{2-8}
                                        & \textbf{Mogo $^{\dagger}$} & \underline{0.505}$^{\pm0.003}$ & 0.693$^{\pm0.003}$ & \underline{0.799}$^{\pm0.003}$ & \underline{0.079}$^{\pm0.002}$ & 3.002$^{\pm0.008}$ & \underline{2.079}$^{\pm0.070}$ 
                                        \\ \hline
    \multirow{9}{*}{\textbf{KIT-ML}}  
                                        & MotionDiffuse\cite{zhang2022motiondiffuse} & 0.417$^{\pm0.004}$ & 0.621$^{\pm0.004}$ & 0.739$^{\pm0.004}$ & 1.954$^{\pm0.062}$ & 2.958$^{\pm0.005}$ & 0.730$^{\pm0.013}$ \\
                                        & T2M-GPT $^{\dagger}$ \cite{zhang2023generating} & 0.416$^{\pm0.006}$ & 0.627$^{\pm0.006}$ & 0.745$^{\pm0.006}$ & 0.514$^{\pm0.029}$ & 3.007$^{\pm0.023}$ & 1.570$^{\pm0.039}$ \\
                                        & Fg-T2M\cite{wang2023fgt2mfinegrainedtextdrivenhuman}& 0.418$^{\pm0.005}$ & 0.626$^{\pm0.004}$ & 0.745$^{\pm0.004}$ & 0.571$^{\pm0.047}$ & 3.114$^{\pm0.015}$ & 1.019$^{\pm0.029}$ \\
                                        & AttT2M $^{\dagger}$\cite{Zhong_2023_ICCV} & 0.413$^{\pm0.006}$ & 0.632$^{\pm0.006}$ & 0.751$^{\pm0.006}$ & 0.870$^{\pm0.039}$ & 3.039$^{\pm0.016}$ & \underline{2.281}$^{\pm0.043}$ \\
                                        & MotionGPT $^{\dagger}$\cite{jiang2024motiongpt} & 0.366$^{\pm0.005}$ & 0.558$^{\pm0.004}$ & 0.680$^{\pm0.005}$ & 0.510$^{\pm0.004}$ & 3.527$^{\pm0.021}$ & \textbf{2.328}$^{\pm0.117}$ \\
                                        & MoMask\cite{guo2023momask} & \textbf{0.433}$^{\pm0.007}$ & \textbf{0.656$^{\pm0.005}$} & \textbf{0.781$^{\pm0.005}$} & \textbf{0.204$^{\pm0.011}$} & \textbf{2.779}$^{\pm0.022}$ & 1.131$^{\pm0.043}$ \\
                                        & MMM \cite{pinyoanuntapong2024mmmgenerativemaskedmotion} & 0.381$^{\pm0.005}$ & 0.590$^{\pm0.006}$ & 0.718$^{\pm0.005}$ & 0.429$^{\pm0.019}$ & 3.146$^{\pm0.019}$ & 1.105$^{\pm0.026}$ \\
                                        \cline{2-8}
                                        & \textbf{Mogo $^{\dagger}$} & \underline{0.420}$^{\pm0.007}$ & \underline{0.634}$^{\pm0.007}$ & \underline{0.754}$^{\pm0.007}$ & \underline{0.313}$^{\pm0.016}$ & \underline{2.957}$^{\pm0.029}$ & 2.063$^{\pm0.066}$ 
                                        \\ \hline
    \multirow{7}{*}{\textbf{CMP} (zero-shot)}  
                                        
                                        & T2M-GPT $^{\dagger}$\cite{zhang2023generating} & 0.061$^{\pm0.003}$ & 0.103$^{\pm0.005}$ & 0.147$^{\pm0.006}$ & 16.092$^{\pm0.099}$ & 4.179$^{\pm0.049}$ & 2.118$^{\pm0.033}$ \\
                                        
                                        & AttT2M $^{\dagger}$\cite{Zhong_2023_ICCV} & 0.065$^{\pm0.004}$ & 0.109$^{\pm0.008}$ & 0.147$^{\pm0.008}$ & 18.403$^{\pm0.071}$ & \underline{4.048}$^{\pm0.017}$ & 2.208$^{\pm0.019}$ \\
                                        & MotionGPT $^{\dagger}$\cite{jiang2024motiongpt} & 0.050$^{\pm0.002}$ & 0.094$^{\pm0.002}$ & 0.133$^{\pm0.003}$ & \textbf{10.654}$^{\pm0.183}$ & 4.431$^{\pm0.021}$ & \textbf{5.535}$^{\pm0.259}$ \\
                                        & MoMask\cite{guo2023momask} & 0.062$^{\pm0.003}$ & 0.108$^{\pm0.005}$ & 0.150$^{\pm0.004}$ & 24.351$^{\pm0.205}$ & 4.817$^{\pm0.022}$ & 1.651$^{\pm0.050}$ \\
                                        & MMM \cite{pinyoanuntapong2024mmmgenerativemaskedmotion} & \underline{0.067}$^{\pm0.004}$ & \underline{0.116}$^{\pm0.008}$ & \underline{0.154}$^{\pm0.008}$ & 17.087$^{\pm0.313}$ & 4.360$^{\pm0.017}$ & 2.802$^{\pm0.011}$ \\
                                        \cline{2-8}
                                        & \textbf{Mogo$^{\dagger}$} & \textbf{0.069}$^{\pm0.003}$ & \textbf{0.119}$^{\pm0.004}$ & \textbf{0.166}$^{\pm0.004}$ & \underline{14.724}$^{\pm0.171}$ & \textbf{4.022}$^{\pm0.029}$ & \underline{3.117}$^{\pm0.066}$
                                         
                                        \\ \hline
    \end{tabular}
  } % End resizebox
  \caption{Comparison with the GPT-type Models of text-conditional motion synthesis on the HumanML3D and KIT-ML test set. $^{\pm}$ indicates a 95\% confidence interval. Bold face indicates the best result, while \underline{underscore} refers to the second best.$^{\dagger}$ denotes a GPT-type model.}
  \label{tab:eval_result_with_GPT}
\end{table*}
This section presents the experimental process and evaluation of Mogo. \cref{sec:exp_data} outlines the datasets and metrics, while \cref{sec:impl_detail} details the training parameters. We discuss optimizing Zero/Few-shot generation via prompt engineering in \cref{sec:infer_optim}, compare our model's results with SOTA motion generation models in \cref{sec:comp_sota}, detail the user study on perceived realism in \cref{sec:user_study}, and explore parameter impacts through ablation experiments in \cref{sec:ablation}.

\subsection{Datasets and Evaluation Metrics}
\label{sec:exp_data}
\textbf{Datasets.} Our model was trained and tested on the HumanML3D \cite{Guo_2022_CVPR} and KIT-ML \cite{Plappert2016} datasets. The HumanML3D dataset includes $14,616$ motion entries from the Amass \cite{Mahmood2019AMASSAO} and HumanAct12 \cite{Guo2020Action2MotionCG} datasets, with a total of $44,970$ textual descriptions (three per entry). The KIT-ML dataset contains $3,911$ motion entries and $6,278$ textual descriptions. We followed the data processing approach of T2M \cite{Guo_2022_CVPR}, splitting the dataset into training, testing, and validation sets in a ratio of $0.8:0.15:0.05$.

CombatMotionProcessed (CMP) dataset \cite{CombatMotion} is a curated collection of 8,700 high-quality motions, all showcasing intense fighting-style motions, with many involving various weapon-based actions sourced from video games. For textual annotations, CMP provides three levels of description for each motion: a concise summary, a sensory-enriched brief, and an extensive, detailed account. As CMP exclusively features non-daily motion types, we leverage its test set to rigorously evaluate our model’s out-of-distribution generation capabilities. This setup offers a precise assessment of generalization performance, underscoring the model’s robustness and adaptability to unfamiliar motion types.
For fairness in evaluating, we do not apply prompt engineering when evaluating Mogo on this dataset.
\\ \textbf{Evaluation Metrics.} We adopted the evaluation framework from T2M\cite{Guo_2022_CVPR}, using metrics such as \textit{Frechet Inception Distance} (FID) for measuring distributional differences between generated and real motions\cite{10.5555/3295222.3295408}. \textit{R-Precision} and \textit{Multimodal Distance} (MM-Dist) assess the consistency of generated motions with input text, where \textit{R-Precision} includes Top-1, Top-2, and Top-3 matching rates. \textit{MultiModality} (MModality) evaluates the variance in distances between multiple motions generated from the same description. We primarily focus on FID to highlight the model's motion quality advantage.

\subsection{Implementation Details}
\label{sec:impl_detail} Our model is implemented using PyTorch. For \textbf{RVQ-VAE}, we use a codebook size of $8192 \times 128$, with a convolution stride of 1 and a dropout rate of 0.2. It is trained on HumanML3D\cite{Guo_2022_CVPR} and KIT-ML\cite{Plappert2016} datasets with 6 quantization layers, using AdamW optimizer and a learning rate of $2 \times 10^{-4}$ over 2000 iterations, with a batch size of 512, trained on an NVIDIA 4090 GPU.

For \textbf{Hierarchical Causal Transformer}, we design 6 sub-Transformers corresponding to RVQ-VAE layers with head counts of $[16, 12, 6, 2, 2, 2]$, layer counts of $[18, 16, 8, 4, 2, 2]$, setting model dimensions to $1024$. We use AdamW optimizer with a CosineAnnealingLR schedule \cite{loshchilov2017sgdr}. For HumanML3D, the learning rate decays from $2.5 \times 10^{-5}$ to $3 \times 10^{-6}$ over $800000$ iterations on an NVIDIA A800-80G GPU with batch size 32 for 1500 epochs. For KIT-ML, it decays from $3 \times 10^{-5}$ to $3 \times 10^{-6}$ on an NVIDIA V100-32G GPU with batch size 48 for 1500 epochs.

\subsection{Inference}
\label{sec:infer_optim}
We use GLM-4\cite{glm2024chatglm} for prompt preprocessing. As shown in \cref{fig:inference}: first, GLM-4 assesses whether the user's prompt requires rewriting by comparing it to text annotations from the training set to determine if it aligns with typical human motions and descriptions. If necessary, GLM-4 reformulates the prompt to match the style of HumanML3D\cite{Guo_2022_CVPR}.The detailed prompt design is shown in the appendix.

On an NVIDIA RTX 4090, the average inference and decoding time per frame is 0.002 seconds. Although the hierarchical structure adds some inference overhead, it still meets the requirements for real-time streaming output in multimedia environments.

% Since this process aims to improve generation for non-standard motions (\eg \textit{"walking like a gorilla" "running like a zombie"}), we disable it during testing, allowing users to toggle the use of the language model for prompt optimization in real-world applications.You can find the detailed prompt design in \cref{sec:prompt_eng_detail}.

\subsection{Comparison to state-of-the-art approaches}
\label{sec:comp_sota}
\begin{figure*}[htbp]
    \centering
    % \captionsetup{justification=raggedright,singlelinecheck=false}
    \includegraphics[width=0.9\linewidth]{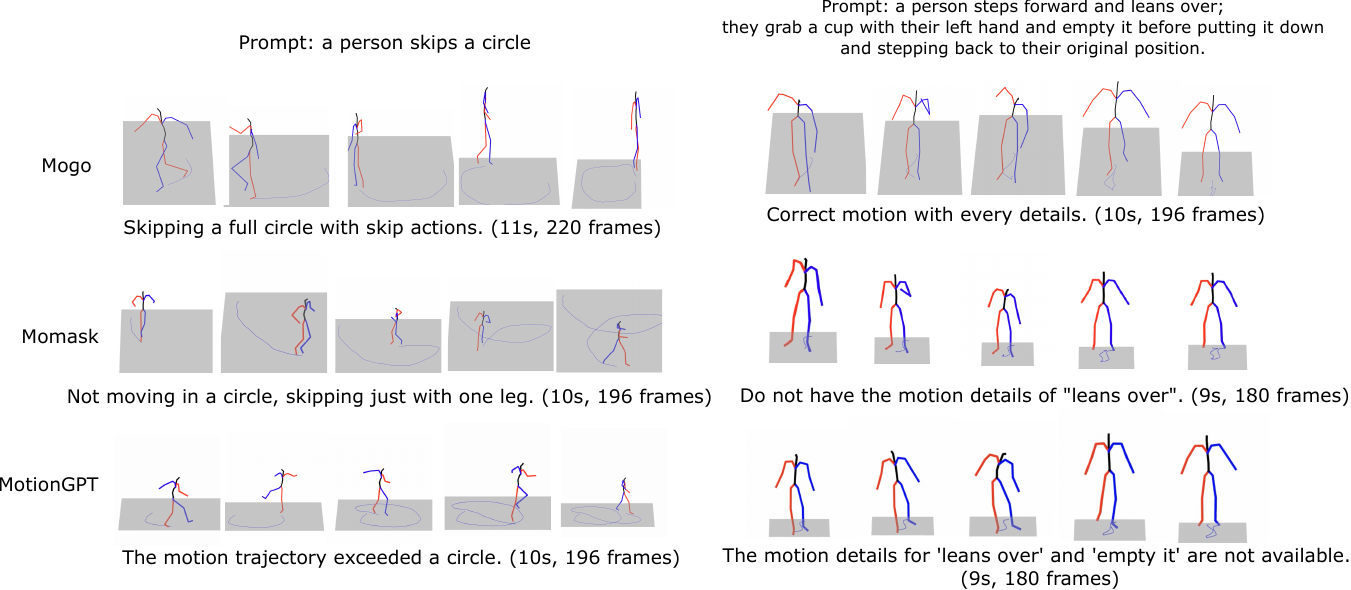}
    \caption{Comparison of the generation quality between our model and the latest SOTA motion generation models\cite{guo2023momask,jiang2024motiongpt}.}
    \label{fig:compare-visual}
\end{figure*}    
We compare our model's performance with state-of-the-art motion generation models through quantitative evaluation and user feedback. \cref{fig:compare-visual} illustrates the visual quality comparison between our model and the latest SOTA models.
\\
\textbf{Quantitative results.} In \cref{tab:compare_vae}, we compare the reconstruction quality of our RVQ-VAE with existing SOTA motion generation models, showing that our model significantly outperforms others in motion reconstruction accuracy. As presented in \cref{tab:eval_result_with_GPT}, our method achieves SOTA performance among all GPT-type models. 
Across all model types, our model ranks second only to MoMask\cite{guo2023momask}. In zero-shot evaluation on the CMP dataset \cite{CombatMotion}, our model achieves SOTA performance, demonstrating superior out-of-distribution generation capabilities. Notably, MoMask\cite{guo2023momask} exhibits a significant drop in all metrics—especially FID—compared to MMM\cite{pinyoanuntapong2024mmmgenerativemaskedmotion} on this dataset, highlighting its susceptibility to overfitting.

    \begin{table}
        \centering
        \resizebox{0.7\linewidth}{!}{ % 调整表格宽度到文本宽度
            \begin{tabular}{ccccc}
            \hline
            Dataset & Methods & FID $\downarrow$ \\
            \hline
                 \multirow{5}{*}{HumanML3D} & TM2T\cite{chuan2022tm2t} & 0.307$^{\pm 0.002}$  \\
                 & M2DM\cite{kong2023prioritycentrichumanmotiongeneration} & 0.063$^{\pm 0.001}$  \\
                 & T2M-GPT\cite{zhang2023generating} & 0.070$^{\pm 0.001}$ \\
                 & MoMask\cite{guo2023momask} & 0.019$^{\pm 0.001}$ \\
                 & MMM\cite{pinyoanuntapong2024mmmgenerativemaskedmotion} & 0.075$^{\pm 0.001}$ \\
                 & \textbf{Mogo} &  \textbf{0.016}$^{\pm 0.001}$ \\
            \hline
            \multirow{4}{*}{KIT-ML}
                & M2DM\cite{kong2023prioritycentrichumanmotiongeneration} & 0.413$^{\pm 0.009}$  \\
                 & T2M-GPT\cite{zhang2023generating} & 0.472$^{\pm 0.011}$ \\
                 & MoMask\cite{guo2023momask} & 0.112$^{\pm 0.002}$ \\
                 & MMM\cite{pinyoanuntapong2024mmmgenerativemaskedmotion} & 0.641$^{\pm 0.014}$ \\
                 & \textbf{Mogo} &  \textbf{0.042}$^{\pm 0.001}$ \\
            \hline
            \end{tabular}
        } % 结束resizebox
        \caption{Comparison of the Reconstruction of our VAE Design vs. Motion VAEs from previous works.}
        \label{tab:compare_vae}
    \end{table}%
    \begin{figure}
        \centering
        \includegraphics[width=0.6\linewidth]{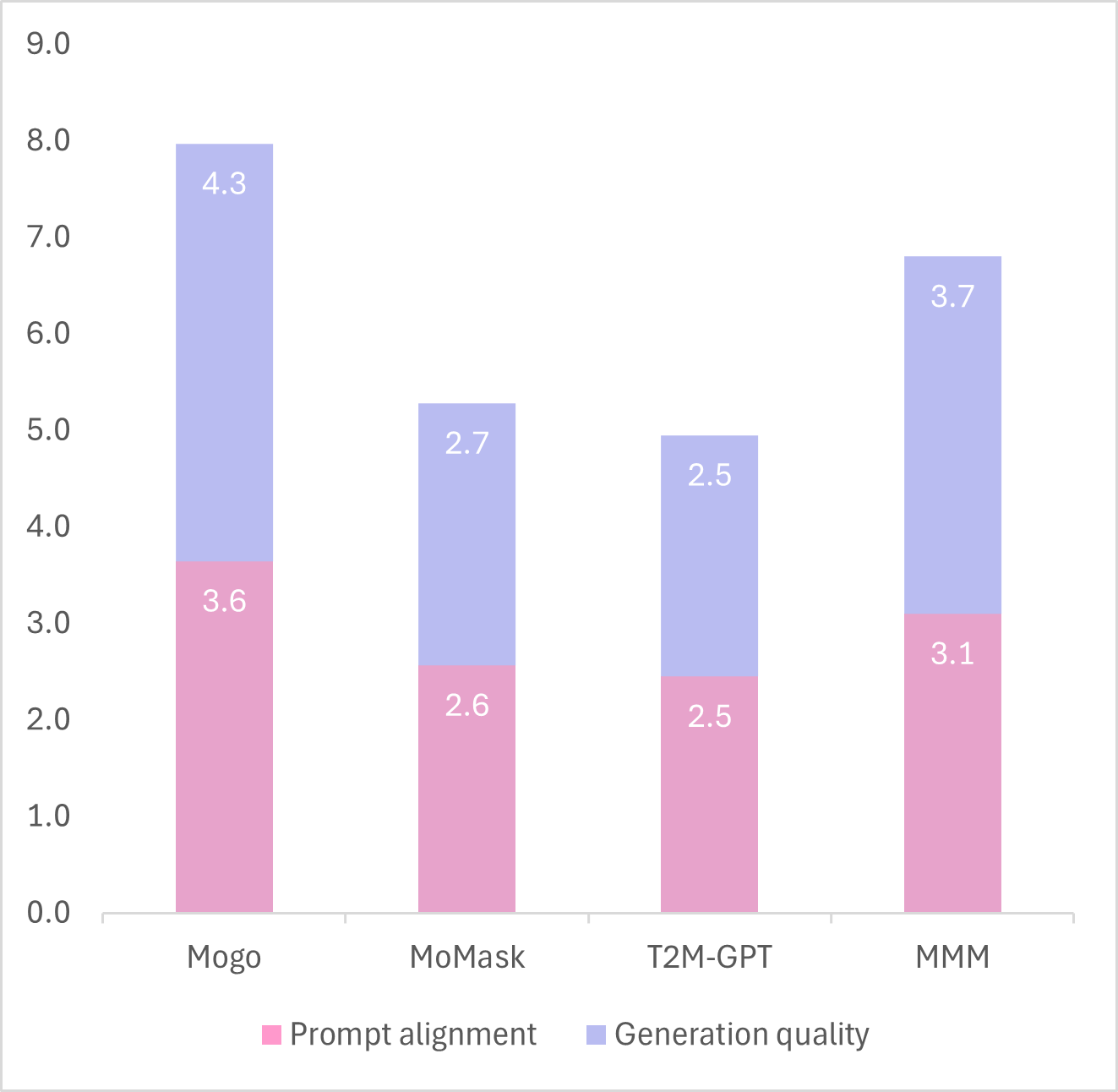}
        \caption{HumanML3D User Study}
        \label{fig:user-study}
    \end{figure}

\begin{table*}[htbp]
    \centering
    \resizebox{0.9\textwidth}{!}{\begin{tabular}{c c c c | c c c}
    \hline
    \multirow{2}{*}{Codebook size}  & \multicolumn{3}{c}{Reconstruction}& \multicolumn{3}{c}{Generation}  \\ \cline{2-7}
                                        & FID$\downarrow$ &Top1$\uparrow$  & MM-Dist$\downarrow$  &  FID$\downarrow$     &   Top1$\uparrow$ &  MM-Dist$\downarrow$  \\     
    % \multirow{2}{*}{\textbf{Codebook size}} & \textbf{FID}$\downarrow$ & \textbf{Top1}$\uparrow$  & \textbf{MM-Dist}$\downarrow$ \\ 
    \hline
    512 $\times$ 512 & 0.022$^{\pm 0.001}$ & 0.508$^{\pm 0.003}$ &  2.997$^{\pm 0.007}$ & 0.203$^{\pm 0.009}$ &  0.469$^{\pm 0.002}$ &  3.138$^{\pm 0.006}$ \\
    1024 $\times$ 1024 & \textbf{0.015}$^{\pm 0.001}$ & 0.511$^{\pm 0.002}$ & 2.984$^{\pm 0.010}$ &  0.184$^{\pm 0.008}$&  0.467$^{\pm 0.003}$&  3.114$^{\pm 0.009}$\\
    2048 $\times$ 512 & 0.017$^{\pm 0.001}$ & 0.511$^{\pm 0.003}$ & 2.980$^{\pm 0.007}$ &  0.092$^{\pm 0.005}$&  0.488$^{\pm 0.002}$&  3.097$^{\pm 0.008}$\\
    4096 $\times$ 256 & 0.019$^{\pm 0.001}$ & 0.510$^{\pm 0.002}$ & 2.989$^{\pm 0.006}$ &  0.090$^{\pm 0.005}$ &  0.491$^{\pm 0.002}$&  3.001$^{\pm 0.008}$\\
    8192 $\times$ 128 & 0.016$^{\pm 0.001}$ & 0.510$^{\pm 0.002}$ & 2.989$^{\pm 0.007}$ &  \textbf{0.079}$^{\pm 0.002}$&  0.505$^{\pm 0.003}$&  3.002$^{\pm 0.008}$\\
    \hline
    \end{tabular}}
    \caption{Study on the number of code in codebook on HumanML3D\cite{Guo_2022_CVPR} test set.Bold face indicates the best FID result}
    \label{ablation:humanml3d}
\end{table*}
\subsection{User Study}
\label{sec:user_study}
Unlike previous studies relying on double-blind user feedback for generation preferences, we conducted a user feedback study using a scoring system focused on \textbf{Prompt alignment} and \textbf{Generation quality}. For \textbf{Prompt alignment}, full motion alignment with the prompt scores $5$ points; partial alignment results in a deduction of $1$ to $3$ points based on motion proportion and order, and no alignment leads to a full $5$-point deduction. For \textbf{Generation quality}, fluent, natural motion earns $5$ points, with $1$ to $2$ points deducted for minor issues and $3$ to $5$ points for major quality problems. We evaluated 25 samples generated by Mogo, MoMask\cite{guo2023momask}, T2M-GPT\cite{zhang2023generating}, and MMM\cite{pinyoanuntapong2024mmmgenerativemaskedmotion} on HumanML3D. \cref{fig:user-study} shows Mogo was significantly preferred by 20 users in a double-blind test.

\subsection{Ablation Study}
\label{sec:ablation}
\textbf{Codebook Size.} As shown in \cref{ablation:humanml3d}, we conducted ablation experiments on the model's reconstruction and generation capabilities on the HumanML3D dataset \cite{Guo_2022_CVPR}, focusing on different codebook sizes. During the experiments, the number of heads for each layer of the Transformer was set to $[16, 12, 6, 2, 2, 2]$, and the number of layers was set to $[18, 16, 8, 4, 2, 2]$. We used the FID of the generated results as the core reference metric, ultimately selecting a codebook size of $8192 \times 128$.
\\
\textbf{Numbers of RVQ-VAE Layers.} In the context of a codebook size of $8192 \times 128$, we investigated the impact of varying the number of layers in the RVQ-VAE on the reconstruction quality. We provide detailed experimental data in the appendix.
\\
\textbf{Impact of dataset size.} As shown in \cref{tab:scaling_data}, we further analyze our model's scaling data capability. We mixed HumanML3D\cite{Guo_2022_CVPR} dataset with different proportions of CMP\cite{CombatMotion} dataset to evaluate the impact of scaling data on model FID performance without retraining the RVQ-VAE, the model's generation quality improves significantly as the data volume increases.
\begin{table}[htbp]
    \centering
    \resizebox{0.9\linewidth}{!}{ % 调整表格宽度到文本宽度
    \begin{tabular}{ccccc}
    \hline
    Epochs  & w/ 0\% CMP  & w/ 50\% CMP & w/ 100\% CMP \\
    \hline
         1 & 24.064 & 22.993  & 18.783 \\
         50 & 0.671 & 0.601 & 0.501 \\
         80 & 0.437 & 0.379 & 0.284 \\
         100 & 0.347 & 0.263 & 0.205\\
    \hline
    \end{tabular}
    } % 结束resizebox
    \caption{The impact of varying dataset sizes on FID performance.}
    \label{tab:scaling_data}
\end{table}
\\
\textbf{Input Condition.} We compared the performance of input conditions by either adding the prompt token and layer token as a sequence prefix or not. As shown in \cref{ablation:compare_cond_res}, our experiments reveal that when without PnQ (adding prompt token to quantization layer token) is applied, the correlation between generated motion sequences and text (measured by R-Precision and MM Distance) decreases significantly. This may result from an additional layer token between action sequence tokens and prompt tokens during training, leading to some attention loss.
\begin{figure}[h]
    \centering
    \begin{subfigure}{0.4\linewidth}
        \centering
        \includegraphics[width=0.4\linewidth]{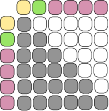}
        \caption{w/o PnQ condition}
        \label{fig:without_add}
    \end{subfigure}
    \hfill
    \begin{subfigure}{0.4\linewidth}
        \centering
        \includegraphics[width=0.4\linewidth]{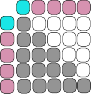}
        \caption{w/ PnQ condition}
        \label{fig:with_add}
    \end{subfigure}
    \caption{Demonstration of Transformer attention masks with different condition structures.}
    \label{fig:compare_cond}
\end{figure}
\begin{table}[htbp]
    \centering
    \resizebox{0.8\linewidth}{!}{\begin{tabular}{c | c c c}
    \hline
    \multirow{2}{*}{Codebook size}  & \multicolumn{3}{c}{w/ PnQ} \\ \cline{2-4}
                                        & FID$\downarrow$ & Top1$\uparrow$  & MM-Dist$\downarrow$  \\     
    \hline
   2048 $\times$ 512 & 0.092$^{\pm 0.005}$ & 0.488$^{\pm 0.002}$ & 3.097$^{\pm 0.008}$ \\
    4096 $\times$ 256 & 0.090$^{\pm 0.005}$ & 0.491$^{\pm 0.002}$ & 3.001$^{\pm 0.008}$ \\
    8192 $\times$ 128 & 0.079$^{\pm 0.002}$ & 0.505$^{\pm 0.003}$ & 3.002$^{\pm 0.008}$ \\
   \hline
    \end{tabular}}
    \resizebox{0.8\linewidth}{!}{\begin{tabular}{c | c c c}
    \hline
    \multirow{2}{*}{Codebook size}  & \multicolumn{3}{c}{w/o PnQ} \\ \cline{2-4}
                                        & FID$\downarrow$ & Top1$\uparrow$  & MM-Dist$\downarrow$  \\     
    \hline
   2048 $\times$ 512 & 0.091$^{\pm 0.005}$&  0.462$^{\pm 0.002}$&  3.293$^{\pm 0.009}$\\
    4096 $\times$ 256 & 0.088$^{\pm 0.005}$ &  0.469$^{\pm 0.002}$&  3.141$^{\pm 0.007}$\\
    8192 $\times$ 128 & 0.083$^{\pm 0.002}$&  0.472$^{\pm 0.003}$&  3.127$^{\pm 0.008}$\\
   \hline
    \end{tabular}}
    \caption{Ablation study on the use of the PnQ condition.}
    \label{ablation:compare_cond_res}
\end{table}
\\
\textbf{Layers and Attention Heads on Model Performance.} 
We tested various configurations of layers and heads for each Transformer in the model on the HumanML3D dataset \cite{Guo_2022_CVPR}. Results in Appendix B indicate that increasing layers and heads enhances generation quality.
\\
\section{Limitations and Discussion}
\textbf{Motion Edit.} Our model is based on a GPT-type architecture with a unidirectional masked autoregressive attention mechanism, lacks the inherent support for temporal completion model editing found in Bert-type models \cite{guo2023momask,pinyoanuntapong2024mmmgenerativemaskedmotion}. This limitation is typical for all GPT-type models.
\\
% \textbf{Zero-shot Inference.}Though prompt engineering greatly reduces out-of-distribution generation failures, some failure cases still arise due to the inability to clearly describe motions. The best solution remains to enhance the dataset's diversity or to fine-tune the model.
% \\
\textbf{Length Limitation of Generation.} Although relative positional encoding enhances Mogo's inference capabilities, allowing it to generate up to 260 frames in continuous, cyclic motion sequences, it is still unable to produce more than 196 frames for non-continuous motions. The lack of inherent correlations in motion data restricts the application of concatenation and the Transformer-XL segment memory mechanism \cite{dai-etal-2019-transformer}. The primary solution to this limitation is to extend the motion sequence lengths in the dataset.
\section{Conclusion}
To leverage GPT-type models' strengths in few-shot learning and streaming output while achieving generation quality on par with or exceeding Bert-type models, we developed Mogo, a text-to-3D motion model using a residual quantizer and hierarchical autoregressive Transformer. Mogo attains SOTA performance across multiple metrics, offering high-quality, extended sequence generation and superior out-of-distribution capabilities.

{
    \small
    \bibliographystyle{ieeenat_fullname}
    \bibliography{main}
}

% WARNING: do not forget to delete the supplementary pages from your submission 
\clearpage
\setcounter{page}{1}
\maketitlesupplementary
\appendix

\section{Overview}
\label{sec:suppl_overview}
In this Appendix, we present:
\begin{itemize}
    % \item Section \ref{sec:compare_all}: Comparison results with all types of SOTA motion generation models.
    \item Section \ref{sec:ablation_archi}: Effects of varying layer numbers and head counts in Transformer layers on HumanML3D generation.
    \item Section \ref{sec:user_study_metrics}: Illustration of the statistical approach applied in the user study.
    \item Section \ref{sec:training_process_data}: Training loss and FID evolution.
    \item Section \ref{sec:prompt_eng_detail}: Prompt engineering details.
    \item Section \ref{sec:length_restriction}: Length restriction.
    \item Section \ref{sec:more_visual}: More visual results of model generations. 
    \item Section \ref{sec:core_code}: Model architecture code.

\end{itemize}
Additionally, we clarify that, to ensure fairness when comparing with other approaches, all evaluation results in the main text are reported without applying prompt engineering optimizations. This decision is based on feedback from real-world applications, where users often provide natural and intuitive prompts such as “walk like a monkey” or “a wizard casting a spell.” Consequently, we treat prompt engineering as a qualitative tool for practical usage rather than a quantitative evaluation strategy.

\section{Ablation study of Mogo architecture}
\label{sec:ablation_archi}
As shown in \cref{tab:compare_vae}, We evaluated the impact of different quantization layers in RVQ-VAE on reconstruction quality.
\begin{table}[htbp]
    \centering
    \resizebox{0.8\linewidth}{!}{ % 调整表格宽度到文本宽度
    \begin{tabular}{ccccc}
    \hline
    Layers  & FID $\downarrow$ & Top 1$\uparrow$ & MM-Dist$\downarrow$ \\
    \hline
         1 & 0.070$^{\pm 0.001}$ & 0.502 $^{\pm 0.001}$ & 2.999 $^{\pm 0.006}$\\
         3 & 0.021$^{\pm 0.001}$  & 0.508 $^{\pm 0.001}$& 2.992 $^{\pm 0.008}$\\
         6 & \textbf{0.016}$^{\pm 0.001}$ & 0.510 $^{\pm 0.001}$ & 2.989 $^{\pm 0.007}$\\
    \hline
    \end{tabular}
    } % 结束resizebox
    \caption{The impact of different quantization layers on model reconstruction quality when the codebook size is $8192 \times 128$. Bold face indicates the best result.}
    \label{tab:compare_vae}
\end{table}
\\
As shown in \cref{tab:ablation_architecture}, we tested the impact of different layer counts and head counts of each Transformer layer in the model on the generation quality using the HumanML3D dataset.the codebook size is $8192 \times 128$.
\begin{itemize}
    \item Parameter A represents heads: \([12, 6, 4, 2, 2, 2]\), layers: \([16, 8, 6, 4, 2, 2]\).
    \item Parameter B represents heads: \([16, 8, 4, 2, 2, 2]\), layers: \([18, 10, 6, 4, 2, 2]\).
    \item Parameter C represents heads: \([16, 12, 4, 2, 2, 2]\), layers: \([18, 16, 6, 4, 2, 2]\).
    \item Parameter D represents heads: \([16, 12, 6, 2, 2, 2]\), layers: \([18, 16, 8, 4, 2, 2]\).
\end{itemize}
\begin{table}[htbp]
    \centering
    \resizebox{0.8\linewidth}{!}{ % 调整表格宽度到文本宽度
    \begin{tabular}{ccccc}
    \hline
    Layers  & FID $\downarrow$ & Top 1$\uparrow$ & MM-Dist$\downarrow$ \\
    \hline
    Parameter A & 0.094$^{\pm 0.003}$  & 0.473 $^{\pm 0.003}$& 3.260 $^{\pm 0.010}$\\
    Parameter B & 0.090$^{\pm 0.003}$  & 0.487 $^{\pm 0.002}$& 3.168 $^{\pm 0.009}$\\
    Parameter C & 0.082$^{\pm 0.004}$ & 0.493 $^{\pm 0.003}$ & 3.063 $^{\pm 0.010}$\\
    Parameter D  & \textbf{0.079}$^{\pm 0.002}$ & \textbf{0.505} $^{\pm 0.003}$& \textbf{3.002}$^{\pm 0.008}$\\
    \hline
    \end{tabular}
    } % 结束resizebox
    \caption{the impact of different layer counts and head counts on the generation results using the HumanML3D dataset. Bold face indicates the best result.}
    \label{tab:ablation_architecture}
\end{table}

\section{User Study Eval Metrics}
\label{sec:user_study_metrics}
\begin{figure}[h]
    \centering
    % \captionsetup{justification=raggedright,singlelinecheck=false}
    \includegraphics[width=0.8\linewidth]{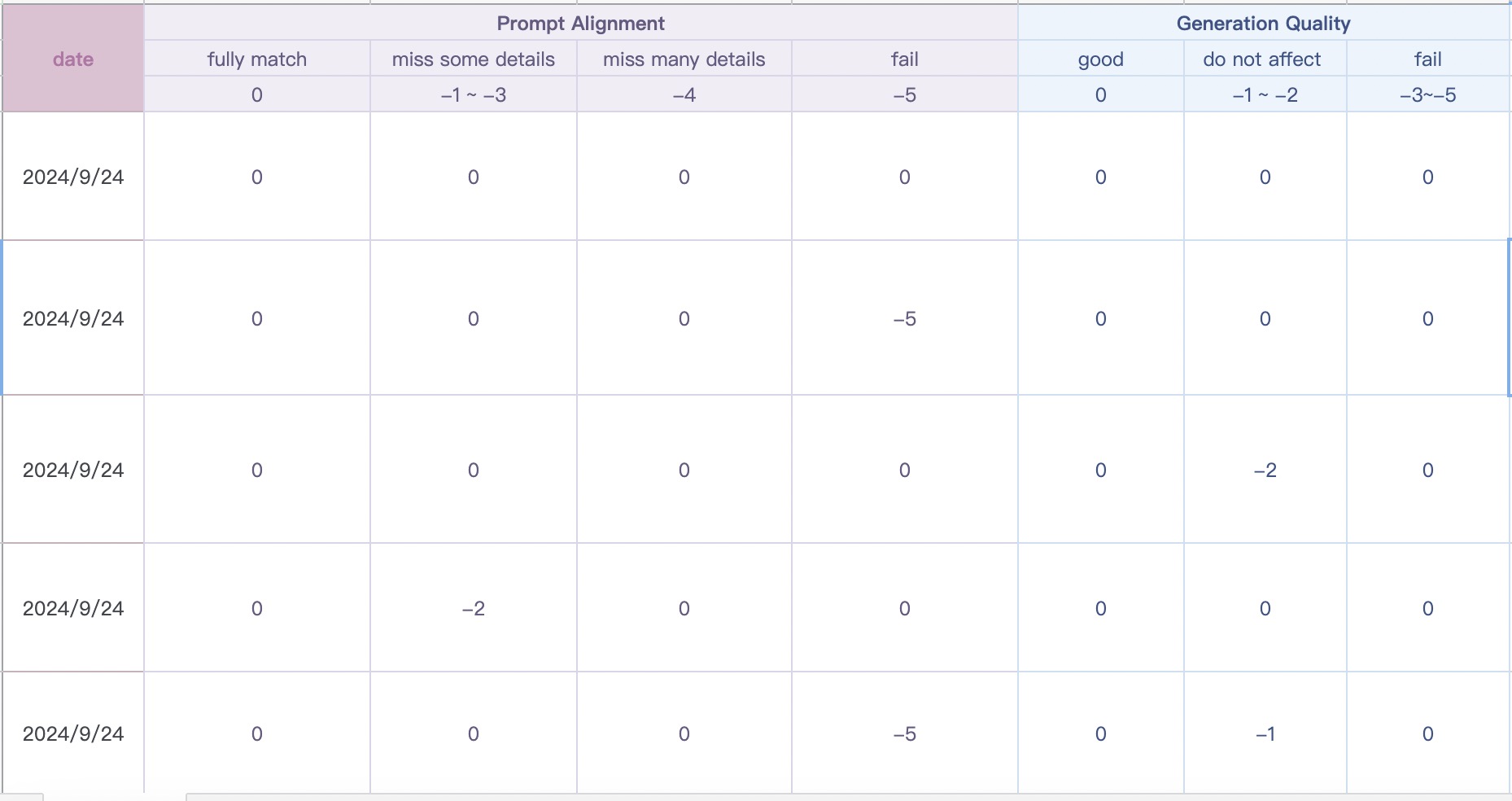}
    \caption{User study eval metrics}
    \label{fig:user-study-metrics}
\end{figure}

\cref{fig:user-study-metrics} illustrate the statistical results of our user study, classified according to the dimensions of experimental evaluation.

\section{Training Loss and FID}
\label{sec:training_process_data}
\begin{figure}[h]
    \centering
    \begin{subfigure}{0.5\textwidth}
        \centering
        \includegraphics[width=\linewidth]{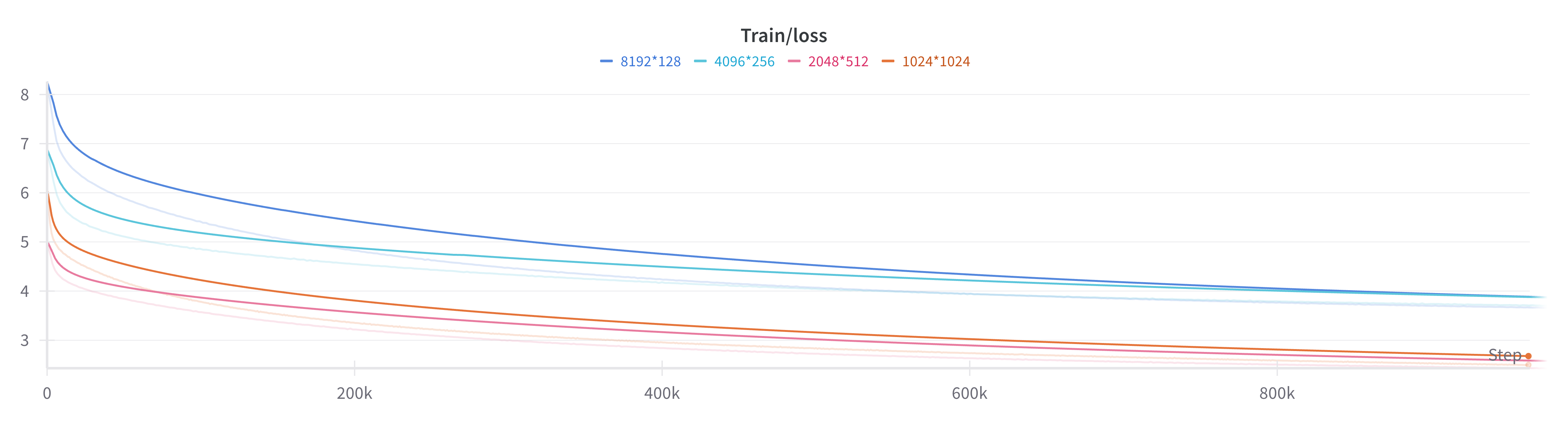}
        \caption{Training Process Loss}
        \label{fig:train_loss}
    \end{subfigure}
    \hfill
    \begin{subfigure}{0.5\textwidth}
        \centering
        \includegraphics[width=\linewidth]{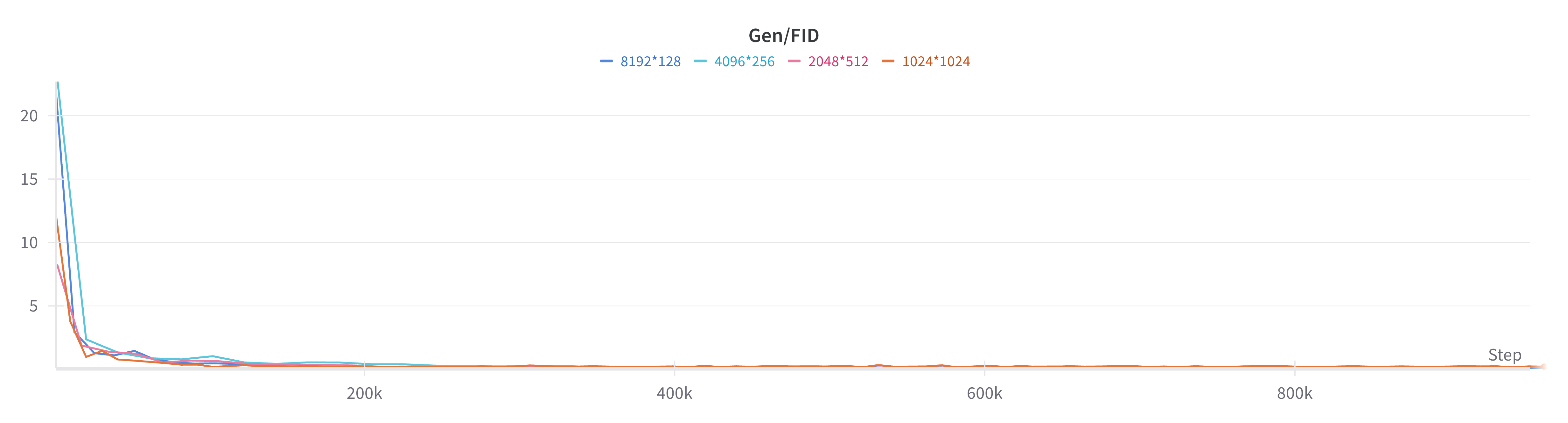}
        \caption{Training Process FID}
        \label{fig:train_fid}
    \end{subfigure}
    \caption{Impact of different codebook sizes on training results in the HumanML3D dataset.}
    \label{fig:training_results}
\end{figure}

\cref{fig:train_loss} and \cref{fig:train_fid} illustrate the effects of different codebook sizes on the loss and generation FID of the Hierarchical Causal Transformer model during training on the HumanML3D dataset(the first 1000 epochs.). The x-axis of both figures represents iterations. During the training process, we evaluate the generation performance on the validation set every 15 epochs.

\section{Prompt Engineering}
\label{sec:prompt_eng_detail}
In this section, I will introduce our prompt engineering cues. The system prompt for the determine phase is:
\\
\textbf{Determine Prompt:}
\begin{quote}
You are now an expert in human motion machine learning. Your task is to determine prompts for a model trained on the HumanML3D dataset, which generates motion sequences from text. I will provide you with some training set prompt examples. Please use these examples to determine whether the user's input needs to be rephrased to better match the dataset's description style. Simply respond with yes or no.
\\
Examples:
\\
person walking with their arms swinging back to front and walking in a general circle\\
a person is standing and then makes a stomping gesture\\
the figure bends down on its hands and knees and then crawls forward\\
a person jumps and then side steps to the left\\
a person casually walks forward\\
The person takes 4 steps backwards.\\
The person was pushed but did not fall.\\
This person kicks with his right leg then jabs several times.\\
a person lifting both arms together in front of them and then lifts them back down
\end{quote}

\textbf{Rewrite Prompt:}
\\
\begin{quote}
You are now an expert in human behavior machine learning. You need to write prompts for a model trained on the HumanML3D dataset that generates motion sequences from text. You need to describe abstract actions directly in English as concrete movements, specifying detailed limb movements and directions. Please output the detailed description directly, limited to one sentence and within 25 words. Do not include interactions with specific objects, only describe human movements. If the input prompt is already a concrete motion description and in English, please return the original input prompt without modification. As a reference, the original dataset contains only everyday actions, boxing actions, and street dance types.
\\
Examples from the training set:
\\
person walking with their arms swinging back to front and walking in a general circle\\
a person is standing and then makes a stomping gesture\\
the figure bends down on its hands and knees and then crawls forward\\
a person jumps and then side steps to the left\\
a person casually walks forward\\
The person takes 4 steps backwards.\\
The person was pushed but did not fall.\\
This person kicks with his right leg then jabs several times.\\
a person lifting both arms together in front of them and then lifts them back down\\
Note:
\\
Do not write specific characters; the action subject should be "a man" or "a person" since there are no specific characters in the training set, such as knights, wizards, soldiers, etc. You should describe their figure through limb movement as much as possible.
Do not include objects being held, as the training set does not have specific objects like swords, knives, or guns. Describe their figure through limb movement instead.
Your description should use simple and clear language, avoiding complex vocabulary.
Try to mimic the wording style of the prompt examples I provided as much as possible.
Examples:
\\
Input: A person anxiously paces after getting up, feeling restless.\\
Output: a man rises from the ground, walks in a circle, and sits back down on the ground.
\\
\\
Input: A medieval knight is fighting.\\
Output: A person stands firmly, raising a sword high, then lunges forward, swinging the sword from right to left while shifting weight onto his front foot.
\\
\\
Input: a man walks in a figure 8\\
Output: a man walks in a figure 8
\\
\\
Input: a man crawls forward\\
Output: a man crawls forward
\\
\\
Input: a person walks in a circle\\
Output: a person walks in a circle
\\
\\
Input: a man is battling\\
Output: a man is boxing and bouncing around
\end{quote}

\section{Length Restriction}
\label{sec:length_restriction}
We did not adopt the length prediction mechanism of GPT-type models, which involves appending an [END] token after each data entry in the dataset to instruct the model when to stop generating. This is because, if we implemented this, the model's generation length would be limited to 196 frames. Instead, we used a length restriction approach, allowing users to input their desired generation length to determine when to stop the generation.

\section{Visualizations of Mogo's Generation}
\label{sec:more_visual}
\begin{figure*}[h]
    \centering
    % \captionsetup{justification=raggedright,singlelinecheck=false}
    \includegraphics[width=\linewidth]{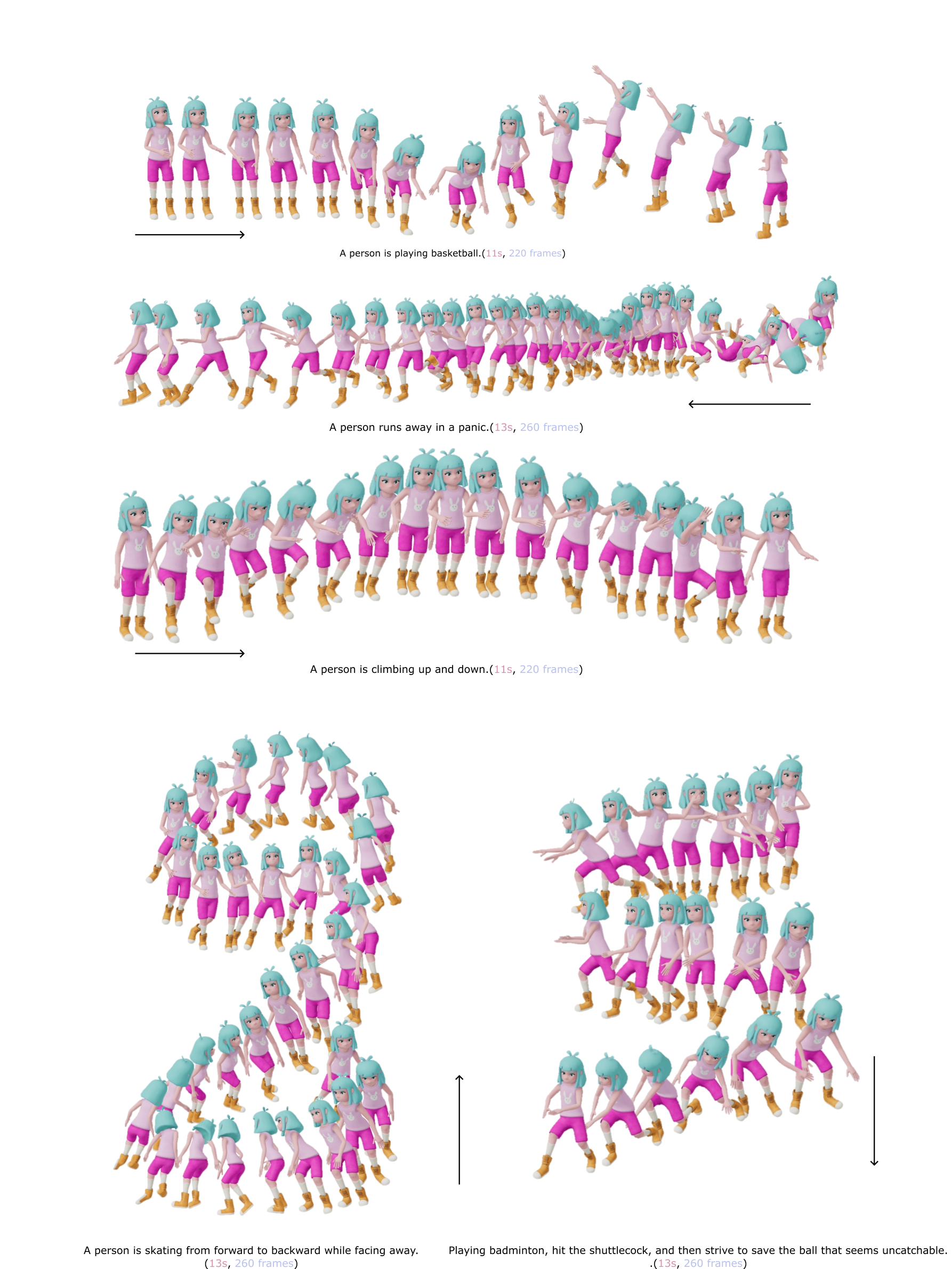}
    \caption{Demonstration of generative capabilities for open vocabulary and ultra-long sequences.}
    \label{fig:more_visual_img}
\end{figure*}

\cref{fig:more_visual_img}  illustrates further exemplary generative capabilities of our model, showcasing its performance in handling complex scenarios, diverse open-vocabulary prompts, and long-sequence generation. The results highlight the model’s robustness in maintaining coherent motion patterns and precise alignment with textual inputs, even in challenging conditions.

\section{Code}
\label{sec:core_code}
The core training code structure of our motion generation model is illustrated below. The variable abbreviations used in the code are defined as follows:\\
\begin{algorithm}
    \caption{Motion Generation Algorithm}
    \begin{algorithmic}[1]
        \State \textbf{Input:} p\_texts, m\_ids, m\_len, lbls
        \State bs, nt, r $\gets$ shape(m\_ids)
        
        \State p\_logits $\gets$ EncodeText(p\_texts)
        \State p\_logits $\gets$ CondEmb(p\_logits)
        \State m\_ids $\gets$ mask\_motion\_token(m\_ids[:, :-1, :])
        \State tks $\gets$ TokEmb(m\_ids)

        \For{i = 0 to n\_q - 1}
            \State r\_tks $\gets$ Reduce(tks, i)
            \State tks\_st[i] $\gets$ r\_tks
        \EndFor

        \State all\_layer\_out $\gets$ []
        
        \For{i = 0 to n\_t - 1}
            \State q\_ids $\gets$ Fill(bs, i)
            \State q\_oh $\gets$ EncodeQuant(q\_ids)
            \State s\_tks $\gets$ QuantEmb(q\_oh)
            \State st\_tks $\gets$ Concatenate(p\_logits + s\_tks, tks\_st[i])
            \State ret, att $\gets$ Transformer(st\_tks, lbls)
            \State layer\_out $\gets$ Head(att)
            \State all\_layer\_out.append(layer\_out)
        \EndFor

        \State \textbf{Output:} out $\gets$ Stack(all\_layer\_out)
        \State ce\_loss, pred\_id, acc $\gets$ CalcPerf(out, lbls, m\_len)
        \State \Return ce\_loss, acc, pred\_id, out
    \end{algorithmic}
\end{algorithm}

\end{document}